---

# ANTI-ROBOT SPECIESISM

---

Julian De Freitas[1*], Noah Castelo[2], Bernd Schmitt[3], Miklos Sarvary[3]

1 – Marketing Unit, Harvard Business School

2 – Marking Unit, University of Alberta

3 – Marketing Division, Columbia Business School

* Corresponding author: jdefreitas@hbs.edu

**Abstract**

Humanoid robots are a form of embodied artificial intelligence (AI) that looks and acts more and more like humans. Powered by generative AI and advances in robotics, humanoid robots can speak and interact with humans rather naturally but are still easily recognizable as robots. But how will we treat humanoids when they seem indistinguishable from humans in appearance and mind? We find a tendency (called "anti-robot" speciesism) to deny such robots humanlike capabilities, driven by motivations to accord members of the human species preferential treatment. Six experiments show that robots are denied humanlike attributes, simply because they are not biological beings and because humans want to avoid feelings of cognitive dissonance when utilizing such robots for unsavory tasks. Thus, people do not rationally attribute capabilities to perfectly humanlike robots but deny them capabilities as it suits them.

In recent years, new artificial intelligence (AI) technologies have been introduced into the marketplace that have the potential to radically change people's work and lives. And now the same generative AI technologies that have enabled intelligent chatbots like ChatGPT are being incorporated into embodied robots (Berkowitz, 2024), enabling robots to perceive, reason about, and physically interact with the world—including with humans[1]. Soon, we may have robots that seem to perfectly resemble us, both in appearance and mind. This paper examines how people might react to robots that seem be "perfectly humanlike".

With major companies like Amazon and Nvidia planning mass production of such robots, we are entering an era where the line between human and non-human entities is increasingly blurred. While technical challenges remain, some existing humanoid robots and digital avatars can already pass as humanlike in certain situations[2]. This research investigates human responses to imagined social interactions with perfectly humanlike humanoid robots. Our findings suggest that the advent of such robots will not lead people to rationally conclude that these robots are as capable as humans in performing some tasks. Rather, people will deny these robots humanlike attributes, driven by their motivation to prioritize their own species and to avoid feelings of cognitive dissonance from utilizing such robots for unsavory tasks.

**Aversion to Robots and AI**

People are often averse to robots. Notably, robots that fall short of resembling humans in subtle ways seem even more eerie or unsettling than robots that are less humanlike—a phenomenon known as the "uncanny valley" (Mori, 1970). Psychological research has explained this effect by arguing that such "almost humanlike" robots appear as aesthetically displeasing, and that they remind people of zombies, death, or disease (Kätsyri et al., 2015; Mori, 1970; Wang et al., 2015). Other psychological explanations focus on how people perceive robot minds, sometimes referred to as the "uncanny valley of mind" (Müller et al., 2021; Stein & Ohler, 2017). These theories suggest that humanoid robots can be unsettling because they remind people of the human ability to experience feelings, even though these robots are not seen as having such capabilities (Gray & Wegner, 2012; Smith et al., 2021). However, these studies are based on today's robots, which in fact *do* still lack some human-level capabilities; thus, these findings may reflect reasonable inferences about true robot limitations.

---

[1] For a demo, see https://www.youtube.com/shorts/bcqn41D5gwI?feature=share; https://www.youtube.com/shorts/vQhqYXGExWY?feature=share

[2] For examples, see https://www.youtube.com/watch?v=uQNtg36cLsg; https://www.youtube.com/shorts/SgE6D--y_3o; https://www.linkedin.com/posts/emollick_i-invited-a-live-heygen-ai-avatar-to-a-zoom-activity-7259367524222328832-g4sa/

A parallel trend to robotics is the advent of generative AI, which has enabled the creation of chatbots that can fool people into thinking they are interacting with a human (Mei et al., 2024; Mitchell, 2024). When people are informed that they are interacting with a chatbot, this triggers "AI aversion"—the notion that people very often prefer to rely on humans rather than AI to perform certain tasks (Castelo, Bos, et al., 2019; De Freitas et al., 2023; Dietvorst et al., 2015), although AI is occasionally preferred (Logg et al., 2019). For example, people generally react more positively to human service providers than AI-powered ones (Castelo, 2023; Luo et al., 2019), unless they are in a setting where it is better to not have a human because humanlike features like social judgment are undesirable, e.g., embarrassing contexts (Holthöwer & van Doorn, 2023; Paluch et al., 2021; Pitardi et al., 2022). More broadly, people's reactions to AI depend on several factors relating to how AI is seen as falling short of human capabilities (e.g., it is seen as emotionless and rigid) impinging upon human capabilities (e.g., when it is very autonomous), and how it is opaque and thus difficult to understand (for a review, see De Freitas et al., 2023).

## The Advent of Perfectly Humanlike Robots

The dual advancements in robotics and AI, combined with strong economic incentives for automating human labor are leading to the advent of a new generation of humanoids that integrate the latest generative AI into their embodiments. More than just a "pretty face," these robots can use AI to reason about and interact with their surroundings (Berkowitz, 2024). With these advancements, we can expect concomitant rises in the demand for robots that are perfectly humanlike. For instance, given that some societies (e.g., Japan) have fast-aging populations, and that 30% to 60% of Americans suffer from chronic loneliness (Beaver, 2021; Holt-Lunstad, 2017), perfectly humanlike robots could offer a technological solution to these issues. Other example applications include customer-facing assistants in hotels, museums, and stores.

How will people respond to perfectly humanlike robots? Will they fully embrace them, or will they still show psychological aversion? Most importantly, what might explain their responses to perfectly humanlike robots?

If the way people view and treat such robots is based on a rational assessment of their capabilities, then we should expect to see people finally treat robots similar to how they treat other human beings; after all, such robots would be externally (in terms of their appearance) and internally (in terms of their mental capacity to think) indistinguishable from humans. On the other hand, if people's views and behaviors are driven more by other motivational factors, then they should deny such capabilities to robots based on whether it is in their interest to do so.

In this paper we explore one such motivational factor—to accord preferential treatment to one's own species. There is some preliminary, suggestive evidence in prior research consistent with such preferential treatment although this research did not include the ultimate test case of perfectly human-like robots. Specifically, robots with anthropomorphic features evoke concerns about human distinctiveness (Ferrari et al., 2016), triggering discomfort from blurring the line between humans and machines; this leads to resistance to using them in service settings (Huang et al., 2021) and reduced support for robotics research (Złotowski et al., 2017). Furthermore, stronger belief in the uniqueness of human nature amplifies these negative outcomes (Ratajczyk et al., 2024).

**Anti-Robot Speciesism**

The term "speciesism" has been used in a psychological sense in the context of human-animal relationships to refer to "the assignment of different moral worth based on species membership (Caviola et al., 2019). Similarly, species-membership may also play a key role in human-robot relations, resulting in "anti-robot" speciesism (Schmitt, 2020). As noted, people seem intent on maintaining a clear boundary between humans and robots and, as a result, they may also deny important attributes to robots that are otherwise indistinguishable from them internally and externally. Specifically, people may "essentialize" differences between humans and robots, such that even when robots perfectly resemble humans they are still viewed as lacking certain inherent, "intangible" properties—ones that are invisible and hard to verify (Bloom, 2005; Gelman, 2003; Medin & Ortony, 1989). As an analogy, people might treat perfect robots much as they do duplicates of original artworks—as lacking certain hidden, essential properties of the original (Hood & Bloom, 2008; Newman et al., 2014). Starting in childhood (Gelman, 2003), people have strong feelings about whether something has essential characteristics of its kind, even if they cannot clearly articulate what exactly those characteristics consist of (Gelman, 2003; Knobe et al., 2013).

One important cue that triggers such essentialist reasoning related to anti-robot speciesism seems to be an entity's biological makeup, since people tend to believe that members of a biological species have a stable, hidden essence that members of that species share with each other but not with other species (Atran, 1993; Gelman & Hirschfeld, 1999; Pinker, 1994). In the case of biological essentialism concerning humans versus other species, people treat the capacity to experience sensations and to have a thinking mind as essential human traits, i.e. ones that are fixed, inborn, and deeply rooted in the person rather than inculcated by a process of learning. Accordingly, across several cultures, people tend to attribute more capacity for experience to themselves than to others (Haslam et al., 2005; Loughnan, Leidner, et al., 2010), and this differential ascription of mind is exacerbated when people reason about out-group members, such as those of different races, ethnicities, or gender (Bain et al., 2012; Costello & Hodson, 2014; Goff et al., 2008; Haslam & Loughnan, 2014; Salmen & Dhont, 2023). Moreover, this prejudice is often driven by the presence or absence of biological markers (Haslam et al., 2000). One way of interpreting such findings is that people view membership in one's group (be it a specific subset of the human species or of the human species overall) as dependent on certain essential traits that are "all or nothing" and immutable (Hirschfeld, 1996; Hull, 1965).

To begin exploring whether similar inferences could be at play when people reason about perfectly humanlike robots, we conducted qualitative interviews in which we found that most participants express a preference for humans over (imagined) perfectly humanlike robots, even when assured these robots would be indistinguishable from humans in every way. When probed for their reasoning, many ultimately cited biology as the key distinction, asserting that being a living organism has inherent value. Participants also believed that biological beings possess intangible qualities that robots fundamentally lack, such as consciousness, a soul, the ability to experience meaning, and the capacity for empathy. These findings suggest that people selectively attribute mental capacities to (biological) humans, while denying them to even perfectly humanlike robots (see Supplemental Material). Building on these qualitative observations, we hypothesize that one cue which humans use to deny perfect robots humanlike capabilities is that, based on knowledge, these robots are not ultimately biological organisms.

**Resolving Cognitive Dissonance from Exploitation of Perfectly Humanlike Robots**

A related, potential driver of the denial of humanlike traits to perfectly humanlike robots could be cognitive dissonance from using robots for unsavory tasks. More Recently, the classic concept of cognitive dissonance has been applied particularly as an explanation in the context of human-animal relationships (Bastian et al., 2012). Human exploitation of animals for eating meat often gives rise to cognitive dissonance, because harm of animals is difficult to justify given that people otherwise love and keep animals as pets, a.k.a the "meat paradox" (Bastian et al., 2012). To reduce this dissonance, people downgrade animals' perceived mental capacities—such as their capacity to suffer, experience emotions, or anticipate the future—thereby justifying harming them (Loughnan, Haslam, et al., 2010) (Bastian et al., 2012). For example, they view kangaroos as less capable of suffering when labeled as "food animals" than when not (Bratanova et al., 2011). In short, people act in ways that allow them to satisfy their own needs while continuing to appear reasonable and moral.

This phenomenon is intimately related to speciesism, since people only exhibit this behavior when judging non-members of the human species. As an informative illustration, people even attribute more moral concern to humans with severe mental disabilities than to more intelligent monkeys, highlighting that their preferential treatment of humans is ultimately based on species membership, not mental capabilities (Caviola et al., 2019).

In sum, people exhibit speciesism, often based on biological or morphological cues. Speciesism is implicated in motivated dehumanization surrounding harm of non-members of one's species, in which the perceived capabilities of these non-members are degraded to alleviate cognitive dissonance. Since perfectly humanlike robots would also be non-members of the human species, they could be subject to the same treatment (Schmitt, 2020).[3] As a crucial departure from previous work that asks participants to evaluate today's participants, however, we focus on such hypothetical robots that perfectly resemble humans, investigating the motivational source of people's evaluations and likely treatment of these robots.

**Dehumanizing Perfectly Humanlike Robots**

If people deny humanlike traits to perfectly humanlike robots, then this would be challenging to justify given the robots' seemingly perfect resemblance in appearance and mind to humans. Under these conditions, we expect that people will still deny robots more "intangible" mental attributes that are harder to observe or verify—such as having a soul, consciousness, or the ability to experience meaning. People have existing intuitions about the existence and nature of such abilities. For example, developmental research finds that from early childhood humans are intuitive dualists, distinguishing between physical bodies and intangible, immaterial souls (Bloom, 2004; Hood et al., 2012) that persist after the death of the physical body (Bering & Bjorklund, 2004). At the same time, such abilities are notoriously difficult for even psychologists and philosophers to verify and agree on.[4] Furthermore, cultures differ on whether non-biological entities can have souls, e.g., cultures that subscribe to Shintoism perceive consciousness/souls in non-biological objects (Castelo & Sarvary, 2022).

---

[3] Note: we do not imply that robots are a "biological species." What is important is that robots are *not* a member of the biological species *homo sapiens,* even though the perfectly human-like robots we study are behaviorally and mentally indistinguishable from humans. We propose that it is this non-membership in the human species that drives anti-robot speciesism. In short, anti-robot speciesism can be understood as a sentiment vis-à-vis robots that treats robots that may have the same skills, capabilities, and appearance differently based on their non-membership in the biological species *homo sapiens.*

For our purposes, it does not matter whether people only dehumanize perfectly humanlike robots for intangible properties rather than tangible properties, although we anticipate that they are more likely to do so for intangible traits and therefore focus on such traits. To assess which qualities are considered tangible or intangible, we conducted a pre-study (N = 98, 46% Male, $M_{age}$ = 42) to measure the perceived tangibility (0 = "Very Intangible", 50 = "Neither Tangible nor Intangible", 100 = "Very Tangible") of eight different qualities. The qualities were inspired both by our qualitative interviews and prior literature on dehumanization of animals (Bastian et al., 2012). We presented them in randomized order on a single page: *consciousness (awareness of oneself and one's surroundings), soul, ability to experience meaning, empathy, growth and evolution over time, comprehension of being harmed, comprehension of wear and tear in one's own body, and contemplation of one's own existence and well-being* (https://aspredicted.org/cksk-jt9v.pdf).

---

[4] For instance, believing in a soul and consciousness is often associated with intuitively subscribing to mind-body dualism, the idea that the mind is separable from the physical brain (Nadelhoffer, 2014), making it effectively impossible to scientifically verify the existence of these constructs. Likewise, one influential philosophical position, known as solipsism, argues that we cannot know for sure that other people have conscious minds; consciousness and other intangible qualities could be just "ghosts in a machine" (Ryle, 1949).

We conducted one-tailed t-tests of whether each quality was lower than the midpoint of the scale. We found that *consciousness, a soul, the ability to experience meaning, empathy, and contemplation of one's own existence and well-being* were all rated as significantly more intangible than tangible (all *ts* < -2.58, all *ps* < .006). Conversely, we failed to reject the null hypothesis for *growth and evolution over time, comprehension of being harmed, and comprehension of wear and tear in one's own body*, indicating that these qualities were not rated as significantly more intangible than tangible (all *ts* > 3.80, all *ps* > .999). Thus, we refer to the first set of traits as "intangible" attributes and the second set of traits as "moderately intangible traits" accordingly, and test whether participants deny each of these types of traits to perfectly humanlike robots in different experiments.

## The Current Studies

Six studies investigate the effects of anti-robot speciesism and cognitive dissonance on comfort, service evaluations, and moral concern toward humanoid robots. Study 1 finds that people feel less comfortable interacting with perfectly humanlike robots than with humans— even though they are more comfortable interacting with such robots than with ones that induce the uncanny valley. Study 2 shows this discomfort is driven by the denial of intangible qualities typically associated with humans. Study 3 demonstrates that beliefs about the lack of a robot's biological origins explain this denial of traits. Study 4 investigates whether framing the robots as capable of consciousness reduces bias, finding increased comfort but persistent concerns about human distinctiveness. Study 5 examines whether initial deception—presenting a robot as human—reduces bias once the robot's true identity is revealed. Study 6 finds that framing robots as workers in hazardous environments leads people to further deny robots the moderately intangible ability to comprehend harm, lowering moral concern towards the robots.

To provide a strong test of our hypotheses, in all studies except study 1, we provided a specific context, in which biological origin should be, by and large, irrelevant. Specifically, rather than using contexts in which biology or evolutionarily conditioned feelings arguably should greatly matter (such as childcare, friendships or counseling), we focused on commercial contexts, presenting robots as employees (e.g. service personnel). We also took several steps to ensure the validity of our stimuli. First, we used a combination of stimuli representing perfectly humanlike robots across our studies, including images, verbal scenarios, and videos of perfectly humanlike robots. Second, we verified that people would indeed view these robots as distinct from existing robots, by comparing them to existing robots. Third, we measured whether people could imagine these robots just as easily as they could imagine human beings—and they could, which is perhaps unsurprising given the public's exposure to advanced robots in the movies and robots. Finally, as a supplementary demonstration to verify some of our statements that future robots that are perfectly humanlike robots could fool humans into thinking they are humans, we demonstrate that even some of today's robots, as well as fully virtual "digital clones" of humans—can already do so, especially under optimal lighting and timing conditions.

**Transparency and openness**

We report how we determined our sample size, all data exclusions (if any), all manipulations, and all measures in the studies. Anonymized trial-level data are publicly available on the Github repository and are accessible at https://github.com/Ethical-Intelligence-Lab/anti_robot_speciesism. Data were analyzed using R version 4.0.2. Two supplemental demonstrations, Study 5, and Study 6 were pre-registered; all other studies were not pre-registered.

## Study 1: First Evidence of Anti-Robot Speciesism

Our first study measures reactions to images depicting humans versus perfectly humanlike robots, to test whether people exhibit more discomfort towards the latter—even though when they would be identical to humans. As a manipulation check of whether such robots are seen as different from today's robots, we also test whether such perfect robots are preferred to typical robots of today.

**Method**

We recruited 204 participants (49% female, mean age = 26) from Prolific Academic. We aimed to recruit 70 participants per condition, which provides 80% power to detect an effect size of $r = 0.48$ or greater in an independent two-sample t-tests comparing mean differences, with a 5% false-positive rate.

Participants were shown either a picture of an imperfectly humanlike robot (i.e., today's robots that tend to elicit the uncanny valley; randomly selected from a set of four robots, to increase generalizability), or a picture of a human being (randomly selected from a set of four humans)—Figure 1. Participants who saw a human were either told that it was a human or an advanced humanoid robot. This experimental design allows us to compare evaluations of perfectly humanlike robots to that for both imperfectly humanlike robots available today and actual humans. To increase the believability of the perfectly humanlike robot manipulation, all participants first watched a brief video showing one of the most humanlike robot heads in existence at the time of the experiment (https://www.youtube.com/watch?v=jTny-MYb4tE).

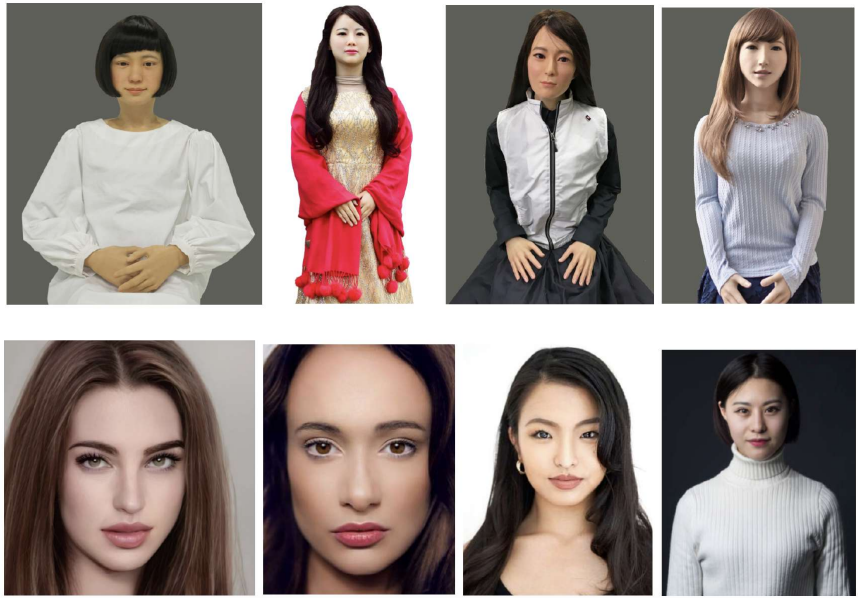

**Figure 1** Stimuli used in Study 1. The top row depicts imperfectly humanlike robots; the bottom row depicts the perfectly humanlike robots / humans.

To measure anticipated comfort interacting with the entity in question, we asked participants how much they would feel "creeped out," "uneasy," and "unnerved" (Gray & Wegner, 2012). These items used 0–10 scales anchored from "not at all" to "completely", and were averaged to create our dependent variable ($\alpha$ = .75). We asked participants whether they had any concerns or thoughts about the survey, planning to exclude anyone who reported suspicion that the robot was in fact a human in the perfectly humanlike robot condition; no participants reported such a suspicion.

**Results**

Participants felt least comfortable with the imperfectly humanlike robots (M = 4.30), relatively more comfortable with the humans portrayed as perfectly humanlike robots (M = 5.67, $t(129)$ = 3.45 $p$ < .001, $d$ = .59), and most comfortable with humans portrayed as such (M = 7.01, $t(133)$ = 3.15, $p$ = .002, $d$ = .54; relative to perfectly humanlike robots). Therefore, while perfectly humanlike robots were seen as a significant improvement over the current state of the art of robots, people still anticipated feeling less comfortable interacting with them than actual human beings. The fact that the identical image elicited lower levels of comfort when participants thought it was a robot versus a human is consistent with anti-robot speciesism; subsequent studies marshal additional evidence.

## Study 2: Denying Perfectly Humanlike Robots Intangible Qualities

In this study, we study the proximal cognitive mechanism underlying lower comfort for perfectly humanlike robots compared to humans—denying such robots the humanlike qualities associated with humans, with a focus on the intangible qualities of having *consciousness and a soul*.

### Method

301 Prolific participants were assigned to one of three conditions. We aimed to recruit 100 participants per condition, which provides 80% power to detect an effect size of $r = 0.40$ or greater in an independent two-sample t-tests comparing mean differences, with a 5% false-positive rate. In the human condition, participants saw a picture of a real human woman and were told this person was an employee at a pharmacy. In the perfectly humanlike robot condition, they saw the same picture of a human, but were told it was an advanced humanoid robot working at a pharmacy. In the imperfectly humanlike robot condition, participants saw a picture of an actual robot (Erica) and were told that the robot works at a pharmacy. The robots were described as able to assist customers, answer questions, and work as cashiers.

Participants then reported how comfortable they would be shopping at a pharmacy that employed this kind of robot (or person). Participants also reported how much the robot or person seemed to have consciousness and a soul. We averaged these items to create a measure of perceived intangible human qualities ($r$ = .86). All measures used 0–10 scales. As in Study 1, this study used an open-ended suspicion probe, asking participants what, if anything, they found suspicious about the study and it coded for whether they mentioned not believing the manipulation. For robustness, we analyzed the results with and without excluding the five participants who did not believe the perfectly humanlike robot was in fact a robot.

**Results and Discussion**

Participants expected to feel least comfortable shopping in a store staffed by an imperfectly humanlike robot (M = 3.92), more comfortable with a perfectly humanlike robot (M = 5.35, $t(205)$ = -3.46, $p$ < .001, $d$ = -.48), and most comfortable with a human (M = 9.11, $t(169)$ = -10.89, $p$ < .001, $d$ = 1.52, compared to a perfectly humanlike robot).

The imperfectly humanlike robot was also seen as having the least intangible qualities (M = 0.90), while the perfectly humanlike robot seemed to have more (M = 1.68, $t(153)$ = 2.91, $p$ = .004, $d$ = .41) and the human had the most (M = 8.60, $t(190)$ = -20.47, $p$ < .001, $d$ = 2.95, compared to the perfectly humanlike robot). Perceptions of intangible qualities were positively related to comfort when looking at all three conditions together ($r$ = .63, $p$ < .001), and when looking only at the two robot conditions ($r$ = .18, $p$ = .009).

To assess whether intangible qualities statistically explain levels of comfort, we used a multi-categorical mediation model with the agent type as the independent variable (treating "human" as the reference level), perception of intangible qualities as mediator), and comfort as the outcome variable. The mediation analysis confirmed that perceptions of intangible qualities mediate the effect of condition on comfort (perfect humanlike robot, indirect effect: $b$ = -2.41, *SE*

= 0.60, 95% CI = -3.52 to -1.17; imperfect humanlike robot, indirect effect: $b$ = -2.17, $SE$ = 0.54, 95% CI = -3.19 to -1.07).

We also reran the analyses excluding the five participants who did not believe the perfectly humanlike robot was in fact a robot and replicated all the results (see Supplemental Material) . In sum, as in Study 1, perfectly humanlike robots were judged as more comfortable to interact with than imperfectly humanlike robots, but less comfortable to interact with than human beings. Consistent with the notion that perfectly humanlike robots are denied humanlike traits, we found that levels of intangible qualities explained these patterns of comfort.

## Study 3: Biology as Antecedent of Dehumanization

This study tests whether the antecedent of ascriptions of humanlike capabilities is intuitions about biological origins, i.e., the notion that perfectly humanlike should be denied traits because they are not biological like humans are.

**Method**

201 Prolific participants (48% female, 3.5% gender non-conforming, mean age = 33.6) were randomly assigned to either the human or the perfectly humanlike robot condition. We aimed to recruit 100 participants per condition, which provides 80% power to detect an effect size of $r$ = 0.40 or greater in an independent two-sample t-tests comparing mean differences, with a 5% false-positive rate. Both conditions showed an image of a human female; in the human condition the female was described as a human while in the perfectly humanlike robot condition she was described as representing a cutting-edge robot that may exist in the near future, followed by the following explanation:

"These robots would be completely indistinguishable from humans in every way -- this is not possible quite yet, but it will be in the very near future. Not only will they look and move exactly like humans, but they will even be able to pass as human when you have long conversations with them. Everything that human employees can do at work, these robots will be able to do as well."

Both the human and the perfectly humanlike robot were described as working in a health-and-wellness store where they would be able to answer questions and provide advice. All participants then answered a series of questions about the person / robot, all on 0–10 scales.

First, we measured service evaluations, asking how comfortable participants would be shopping in a store where the human/robot is employed, how likely they would be to visit such a store, and whether they thought the human/robot would be able to provide effective customer service. These three items were averaged to form our dependent variable (α = .87). Next, we measured the degree to which the human/robot was seen as a biological organism, asking to what extent it was a "living organism" and a "biological being" (α = .99). Finally, we asked about perceived intangible qualities, specifically about having "consciousness," "a soul," "empathy," and the ability to "experience meaning" (α = .96).

**Results and Discussion**

Service evaluations were less positive in the perfectly humanlike robot condition (M = 5.27) than in the human condition (M = 6.83, $t(190)$ = -4.88, $p$ < .001, $d$ = -.69). Furthermore, the human seemed like more of a living organism (M = 7.62) than the perfectly humanlike robot (M = 0.95, $t(141)$ = 18.23, $p$ < .001, $d$ = 2.58) and seemed to have more intangible human qualities (M = 7.01 vs. 1.72, $t(159)$ = 15.20, $p$ < .001, $d$ = 2.17).

To test the full proposed psychological mechanism, we used a serial mediation model, in which people's biological intuitions lead to a higher perception of intangible qualities in humans compared to humanlike robots (condition → biological origin → intangible qualities → service evaluations). Supporting this model, we found that perceptions of the entity as non-living organism predicted perceived lack of intangible qualities ($b = 0.80$, $p < .001$), which in turn predicted service evaluations ($b = 0.29$, $p < .001$), with the full model showing a significant indirect effect ($b = 1.07$, 95% CI = .03 to 2.33). We also note that the entity condition predicted service evaluations when *not* controlling for the two serial mediators ($b = -1.56$, $p < .001$), but did not predict service evaluations controlling for them ($b = 0.19$, $p = .706$). We also note that judgments about biology directly predicted service evaluations ($b = 0.24$, $p < .001$), further suggesting that the effect is driven by intuitions about biology. Altogether, these results support the interpretation that the reason perfectly humanlike robots are denied intangible qualities and evaluated less positively is that they are viewed as non-biological entities.

## Study 4: Manipulating Intangible Qualities

While it would be infeasible to change people's beliefs about the biological origins of perfectly humanlike robots, is it possible to improve attitudes toward such robots by convincing people that they could, in fact, have intangible qualities? In Study 4, participants were exposed to a message by an expert, providing arguments that robots either can or cannot have a conscious mind. These arguments were drawn directly from the philosophical literature on mind and consciousness—specifically from the "dualism vs. physicalism" debate that portrays the human mind as having non-material aspects that cannot be replicated in a machine ("dualism") or as something explainable entirely in terms of physical brain processes that can be replicated in a machine ("physicalism") (Dennett 1997).

### Method

200 Prolific participants (49% female, mean age = 26) were randomly assigned to watch one of two videos, featuring a professor explaining that robots either could or could not have a conscious mind in the same way that humans do (see

https://www.youtube.com/watch?v=SI4xaihN8Nk and https://youtu.be/QvWQafW3NWg). We

aimed to recruit 100 participants per condition, which provides 80% power to detect an effect size of $r = 0.40$ or greater in an independent two-sample t-tests comparing mean differences, with a 5% false-positive rate. This manipulation aimed to convince participants that, in fact,

perfectly human robots could (versus could not) have intangible qualities typically deemed as

essential to humans, including a conscious mind. We refer to these conditions as the "mind" and

"no mind" conditions. After watching their assigned video, participants were asked to write

about why the arguments in the video were likely true—a technique known as the "saying-is-

believing technique," commonly used to increase engagement with stimuli and facilitate attitude

change (Higgins and Rholes 1978). As a manipulation check, we asked participants whether they

believed that robots could eventually have conscious minds, on a 0 (not at all) to 10 (completely)

scale.

      All participants were then shown a picture of a humanoid robot, Erica. The picture was

accompanied by an explanation that humanoid robots are approaching perfect humanlikeness and

are being used as employees in stores, restaurants, and hotels, and that one such robot had even

been granted citizenship in Saudi Arabia. We asked participants how comfortable they would be

(a) shopping in a store and (b) dining in a restaurant where this kind of robot was employed, and

(c) how interested they would be in having the robot as a social companion. Next, because

people who are more comfortable with robots may be more willing to promote their

development, we also told participants that we would be donating $1 on behalf of each

participant to an organization working on human-robot relations and asked them to decide which organization we would donate to on their behalf. They were given a choice between the American Society for the Prevention of Cruelty to Robots, which was described as working to advance the development of human-like robots, and the Center for the Study of Existential Risk, which was described as working to prevent the development of human-like robots. Both organizations are real.

We also asked participants how much this kind of robot seemed to threaten human jobs, threaten human safety, and how much it seemed to blur the line between humans and machines (i.e., a threat to human distinctiveness). Finally, we asked how engaging and convincing the video was and how knowledgeable the speaker seemed. All of these measures used 0 (not at all) to 10 (completely) scales.

**Results and Discussion**

The two videos were judged to be equally engaging, but the speaker seemed more convincing and knowledgeable in the "no mind" video (see Table 1 for full results). The manipulation check showed that the videos successfully altered the belief that robots could have a conscious mind. Note that although it was unintended that the "no mind" video was more convincing, this in fact makes our test more conservative, because we would have had a better chance of showing that the "mind" manipulation is effective (as hypothesized) if the "no mind" video had been *less* convincing.

People expected to be more comfortable in a store and a restaurant staffed by robots and were more interested in having a robot as a social companion in the mind condition, when they believed that robots could have a conscious mind. Reflecting this increased comfort, participants in the mind condition were also more likely to choose the pro-robot organization for their donation and evaluated companies employing robots more positively—Table 1. Note that these results are significant for all measures except the pro-robot donation, although results for this measure were in the predicted direction. In addition, while robots in the two conditions were deemed to pose similar threats to human safety or jobs, robots in the mind condition posed a greater threat to human distinctiveness (see last three measures in Table 2). The fact that participants were more comfortable with robots in the mind condition despite this increased distinctiveness threat indicates that humans have a sufficiently strong preference for robots with intangible human qualities like consciousness and a soul that it outweighs this threat.

**Table 1**. Results of Study 4. All scales were 0–10 except donation choice, which was binary.

|  | **No Mind** | **Mind** | **t-test** |
|---|---|---|---|
| **Engaging** | 5.82 | 5.81 | $t = .03, p = .978$ |
| **Convincing** | 6.82 | 5.82 | $t = 3.02, p = .003$ |
| **Knowledgeable** | 7.62 | 6.79 | $t = 2.80, p = .006$ |
| **Mind (MC)** | 3.02 | 6.36 | $t = -8.90, p < .001$ |
| **Store Comfort** | 3.88 | 5.11 | $t = -3.18, p = .002$ |
| **Restaurant Comfort** | 3.71 | 5.19 | $t = -3.63, p < .001$ |
| **Social Companion** | 1.87 | 3.50 | $t = -4.31, p < .001$ |
| **Pro-robot Donation** | 40% | 50% | $\chi^2 = 1.64, p = .201$ |
| **Safety Threat** | 4.33 | 4.21 | $t = .30, p = .767$ |
| **Job Threat** | 6.35 | 6.09 | $t = .63, p = .526$ |
| **Distinctiveness Threat** | 4.08 | 5.06 | $t = -2.44, p = .016$ |

These results provide support for part of our theorizing—that perfectly humanlike robots would be treated differently because of whether they would be accorded humanlike traits—since convincing people that such robots could have these traits improves attitudes toward them.

## Study 5: Blurring Human-Robot Category Distinctions And Trait Speciesism Moderator

Participants in the previous studies seem to assume, by *default*, that there is a fundamental difference between the "human" and "robot" category, rooted in biology. In Study 5 we explore whether first blurring the human-robot category distinction affects anti-robot speciesism. To do so, we make people initially hold the mistaken belief that they are interacting with a human service agent, only to later learn that it was actually a robot. We reasoned that it might be possible that people change their perception of the category, "robots", if they are subsequently told that the interaction that they believed to be with a "human" was, in fact, with a "robot." Alternatively, people may discount this experience, reverting to anti-robot speciesism. As a further test of the anti-robot speciesism interpretation, we also measure individual differences measure in trait-level speciesism, testing if it moderates the denial of humanlike traits. Finally, as a stimulus check, we test whether perfectly humanlike robots are just as easy to imagine as humans.

### Method

We recruited 900 participants from Prolific and asked them to first complete a measure of trait-level anti-robot speciesism ($\alpha = .74$). The measure was adapted from a speciesism scale developed for measuring speciesism towards animals (Caviola, Everett, and Faber 2019), e.g., by replacing the terms "animals" with "robots": "Morally, robots will always count for less than humans"; "Humans should always have the right to use robots however they want to"; "It should always be morally acceptable to trade robots like possessions"; "If robots become indistinguishable from humans, they should have basic legal rights such as a right to life or a prohibition of torture." To reduce the chance of spillover effects in which measuring this variable affects responses to the dependent variables, we measured it two days prior to the main survey. 688 participants from the first survey spontaneously took the main survey two days later. Note that this is a typical attrition rate, in that not all participants who are eligible for a main survey two days later may choose to take it. Some participants did not answer all questions; our final sample after removing missing cases is therefore 610. We aimed to recruit 200 participants per condition, which provides 80% power to detect an effect size of $r = 0.28$ or greater in an independent two-sample t-tests comparing mean differences, with a 5% false-positive rate.

Participants were then assigned to one of three conditions. The human and perfectly humanlike robot conditions paralleled previous studies. Participants in the third condition (the "surprise robot" condition) were first told that the picture showed a human employee and answered the same series of questions as in the other two conditions. After completing those questions, however, participants in this condition were then informed that the presumed human was in fact a perfectly humanlike robot. These participants then completed the same series of questions a second time, considering that they were now evaluating a robot and not a human as they initially believed.

Our primary dependent variable was a composite measure of service evaluation including participants' anticipated comfort interacting with the entity, their stated likelihood of visiting a store where the entity is employed, and how likely they would be to recommend such a store to their friends ($\alpha = .91$). Participants in the perfectly humanlike robot and human conditions only responded to these questions once, while those in the surprise robot condition responded to them twice (before and after finding out that they had interacted with a robot). We use the second set of responses from participants in this condition, since those reflect their perceptions after learning that they had interacted with a robot. To test whether participants had difficulty imagining the kinds of robots we described, we also asked them how easy it was to imagine the robot (or person) described, as well as how confident they were in their responses to the questions about the robot (or person). All questions in this study used 0–10 scales. This study was pre-registered at https://aspredicted.org/72N_YNN.

**Results and Discussion**

As expected, participants had no more difficulty imagining the perfectly humanlike robot that we described (M = 7.71) than the person that we described (M = 7.86, $t(409)$ = 0.84, $p$ = .402, $d$ = 0.08). The surprise robot was also equally easy to imagine (M = 7.91, $t(402)$ = 0.79, $p$ = .785, $d$ = 0.03, compared to the human).

Replicating the prior studies, participants' service evaluations were more negative in the perfectly humanlike robot condition (M = 5.67) than in the human condition (M = 6.44, $t(379)$ = 3.03, $p$ = .003, $d$ = 0.30). Evaluations were also more negative in the surprise robot condition (M = 5.53, $t(377)$ = 3.65, $p$ < .001, $d$ = 0.36, compared to the human condition; $t(401)$ = 0.47, $p$ = .641, $d$ = 0.05, compared to the perfectly humanlike robot condition). Learning that they had been interacting with a robot *after* initially evaluating it as a human therefore did not seem to affect participants' responses, relative to knowing the robot's identity from the outset. In short, as soon as people believe they are interacting with a robot, they seem to revert to evaluating it more negatively?

Do these effects depend on participants' trait-level speciesism? We conducted a regression analysis to test the effects of condition, trait-level speciesism, and their interaction on service evaluations. The perfectly humanlike robot and surprise robot conditions were dummy coded (0 / 1) and the human condition was set as the reference group. The only significant effect in this regression was the interaction between the perfectly humanlike robot dummy variable and trait speciesism, $b = -.43$, $p = .025$. We used a floodlight analysis to identify how the effects of condition vary depending on participants' trait level of speciesism. We found that among participants with relatively high trait speciesism (above 4.4 on the 11-point scale, or 65% of the sample), service evaluations were more negative in the perfectly humanlike robot condition than in the human condition, $b = -1.27$, $p < .001$. However, among participants with relatively low trait speciesism (below 4.4, or 35% of the sample), service evaluations were not significantly different between the perfectly human-like robot and human conditions, $b = 0.17$, $p = .660$.

Trait speciesism did not interact with the surprise robot dummy variable ($b = -0.06$, $p = .738$), suggesting that evaluations in this condition were uniformly negative across the distribution of trait speciesism. This may be due to perceived deception or an otherwise unwelcome surprise in this condition, such that even participants low in trait speciesism still react negatively when they are only informed about the robots' identity after interacting with it. Nevertheless, the interaction pattern in the perfectly humanlike robot condition is consistent with our theorizing and suggests that baseline levels of trait speciesism matter in shaping reactions to humanoid robots.

**Study 6: Motivated Dehumanization of Robots Used In Dangerous Conditions**

This final study tests whether people would also dehumanize perfectly humanlike robots to avoid cognitive dissonance from using these robots for unsavory purposes. We predict that framing robots as dying from being used in dangerous conditions will lead people to deny these robots the (moderately intangible) ability to comprehend that they are being harmed, which in turn will result in lower moral concern compared to a control condition wherein the robots are merely described as mentally and physically indistinguishable from humans. To disentangle such an effect from the mere mention of robot mortality, we also include a "death-control" condition (inspired by Bastian et al., 2012), in which the robots are described as naturally deteriorating over time rather than dying because of the hazardous work itself. We predict that perceptions in the death-control condition will not differ from those in the control condition, ruling out the mere salience of robot mortality as the underlying cause of robot dehumanization. Finally, as a further generalization test, the current study utilizes video stimuli of robots rather than images.

**Method**

We recruited 600 participants from Prolific and randomly assigned them to one of three experimental conditions (control, death-control, hazardous-work) in a between-subjects design. We aimed to recruit 200 participants per condition, which provides 80% power to detect an effect size of $r = 0.28$ or greater in an independent two-sample t-tests comparing mean differences, with a 5% false-positive rate. 81 participants failed the comprehension check, leaving us with a final sample of 519. Participants were then introduced to the scenario and asked to imagine the very near future, where humans have created a robot that is indistinguishable from humans both physically and mentally. The scenario stated:

> "Humans have created a robot that is indistinguishable from humans, both physically and mentally.
>
> Not only will these robots look and move exactly like humans, but they will also be able to pass as human during long conversations. Everything humans can do, these robots can do as well."

Participants then watched a video to enhance the realism of the scenario, in which a supposed perfectly humanlike robot introduced itself (https://youtu.be/unFy7SCEwxk). The video was created using HeyGen, an AI-powered platform that generates photorealistic avatars. The avatar was AI-generated, with the software syncing its movements, facial expressions, and voice to match the provided script, resulting in a highly humanlike video.

Participants were then randomly assigned to one of three conditions. In all conditions, participants were informed that the robots are completely indistinguishable from humans, both physically and mentally, stating that "People who have interacted with the robot say they couldn't tell it wasn't human because it acts just like a person and holds conversations that feel completely natural." In the control condition, this was the only information provided to the participants. In the hazardous-work condition, the robots were additionally framed as being designed to work in hazardous radioactive mines, where they ultimately die:

> "These robots are sent to work in environments too hazardous for humans, such as in radioactive mines. Due to the high levels of radiation, the robots experience damage, which leads to a premature and inevitable death."

In the death-control condition, the robots were instead framed as naturally deteriorating over time rather than being actively harmed in hazardous work:

> "These robots are designed to function for extended periods. However, like humans, their components deteriorate and malfunction over time. Eventually, critical components fail, leading to their eventual death."

On the next page, participants rated the extent to which they agreed with the following randomized statements on their perception of the robots' capacity to comprehend harm (0 = "Strongly Disagree", 100 = "Strongly Agree"): "The robot would comprehend the existential significance of wear and tear in its body parts."; "The robot would have the capacity to understand when it is being harmed."; and "The robot would have the capacity to contemplate its own existence and well-being." These items were averaged to create a composite measure of the perceived cognitive capacity to comprehend harm ($\alpha = .89$). The statements were adapted from measures on the capacity to suffer in animals in Bratanova et al. (2011).

On the same page, participants rated the extent to which they agreed with several randomized statements on their moral concern for the robots on 100-point scales (0 = "Strongly

Disagree", 100 = "Strongly Agree"), adapted from (Bastian et al., 2012): "It would be morally wrong to disregard the robot's autonomy or force it into decisions against its will."; "It would be morally wrong to intentionally harm the robot when it can be avoided."; "It would be morally wrong to let the robot die when it can be avoided."; "It would be morally wrong to support and engage in practices that exploit the robot for personal gain." These items were averaged to form a composite measure of moral concern (α = .89).

Participants then completed a comprehension check to ensure they understood the scenario they were assigned, in which robots were either only perfectly indistinguishable from humans (control), deteriorated naturally (death-control), or died from hazardous work environments (hazardous-work). Those who failed were excluded from the analysis. This study was pre-registered at https://aspredicted.org/7thq-9626.pdf.

**Results and Discussion**

*Perceived Cognitive Capacity to Comprehend Harm.* A univariate ANOVA was conducted with experimental condition (control, death-control, hazardous-work) as the independent variable and the perceived cognitive capacity to comprehend harm as the dependent variable. We found a significant main effect of condition on perceived cognitive capacity to comprehend harm, ($F(2, 516) = 4.58$, $p < .011$, $\eta^2 = .02$). Pairwise t-test comparisons indicated that robots in the hazardous-work condition were rated significantly lower in this capacity (M = 40.05) than those in the control condition (M = 47.21, $t(336) = 2.30$, $p = .022$, $d = 0.25$) and the death-control condition (M = 48.58, $t(332) = 2.79$, $p = .006$, $d = 0.30$). There was no significant difference between the control and death-control conditions ($p = .633$).

*Moral Concern.* A second univariate ANOVA was conducted with moral concern as the dependent variable, and we found a significant main effect of condition on moral concern, ($F(2,$

516) = 16.75, $p < .001$, $\eta^2 = .06$). Pairwise t-test comparisons indicated that robots in the hazardous-work condition were rated significantly lower in moral concern (M = 35.32) than those in the control condition (M = 51.14, $t(340) = 5.28$, $p < .001$, $d = 0.57$) and the death-control condition (M = 49.49, $t(340) = 4.65$, $p < .001$, $d = 0.50$). There was no significant difference between the control and death-control conditions ($p = .574$).

A multi-categorical mediation analysis with the hazardous-work condition as the baseline confirmed that lower perceived cognitive capacity to comprehend harm in the hazardous-work condition mediated the reduction in moral concern (control indirect effect: $b = 4.17$, $SE = 1.83$, 95% CI = 0.61 to 7.77; death-control indirect effect: $b = 4.96$, $SE = 1.80$, 95% CI = 1.47 to 8.49).

These results support the hypothesis that framing robots as working in hazardous environments lowers leads people to deny them the mental capacity to comprehend harm, which in turn reduces moral concern toward them. More broadly, people appear to dehumanize even perfectly humanlike robots when they need to alleviate uncomfortable feelings in using robots for unsavory purposes.

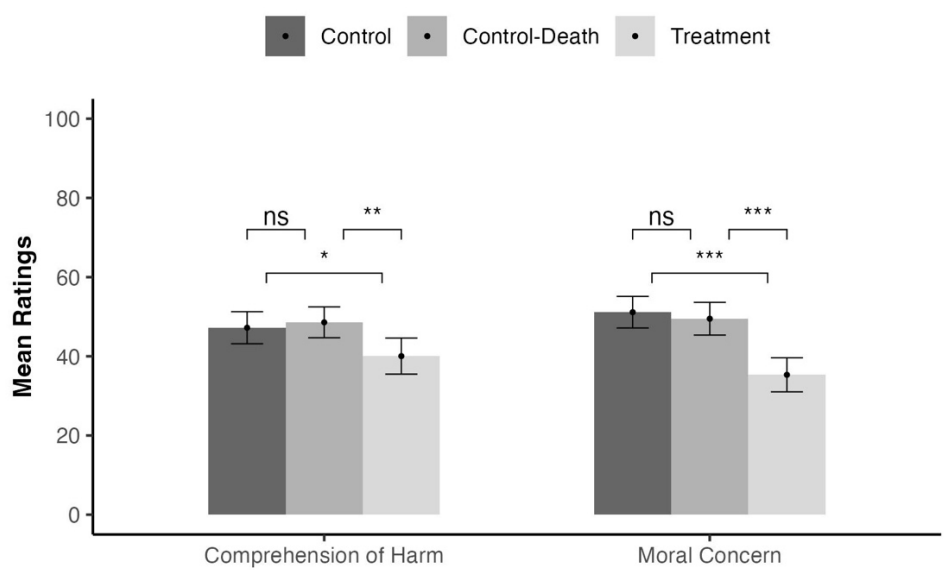

**Figure 3.** Effect of Framing Robots as Dying from Being Used in Hazardous Work on Comprehension of Harm and Moral Concern (Study 6).
Note: ns *p* > .05, * *p* < .05, ** *p* < .01, *** *p* < .001

## GENERAL DISCUSSION

In six experiments we showed that perfectly human-like robots are treated differently from humans because of a tendency that we call "anti-robot speciesism." Because robots are not biological organisms, they are viewed as lacking human qualities deemed essential to being a human, such as consciousness and a soul—even when they are indistinguishable from humans. These beliefs ultimately rest on the notion that "substrates matter," such that if a carbon-based biological organism claims to have consciousness, that claim is accepted, but not so for a silicon-based robot because of the substrate comprising the entity. The results are robust across several conditions. In the Supplemental Material, we report two additional robustness experiments, which show that anti-robot speciesism persists even when robots are utilized in purely mechanical (non-people-facing tasks) and even when participants are reminded of how robot's unique capabilities (e.g., superior memory) would make them highly beneficial in service settings.

### Theoretical Contributions

First, we contribute to work on resistance to AI and robots. Previous work makes diverging predictions about how people will react to perfectly humanlike robots. One natural assumption is that resistance to humanoids will be alleviated once they are indistinguishable from humans, because they will be equally capable. Likewise, work on the uncanny valley suggests that robots may overcome eerie associations once they perfectly resemble humans rather than fall slightly short of perfect resemblance (Mori, 1970). We find that perfectly humanlike robots would be viewed more favorably than uncanny robots, yet remain less preferred than humans. We provide an account, anti-robot speciesism, for why such an effect might ultimately arise. Indeed, it is possible that the same mechanism uncovered here is also at play in people's lower preferences of robots that are truly less capable than humans, leading them to attribute even such (less capable) robots fewer capabilities than they truly have.

Second, we contribute to psychological work on dehumanization, which has suggested that people may dehumanize robots much as they dehumanize outgroups and women (Gray & Wegner, 2012; Smith et al., 2021). Yet these works have studied today's robots, which people may reasonably judge as less capable based on their beliefs of current-day robot capabilities. By holding these capabilities constant between robots and humans, as well as measuring the role of individual differences in trait speciesism and framing of robots for unsavory purposes, we provide a clear demonstration of dehumanization of robots (via the denial of intangible attributes to them and lower preference for interacting with them) and show how this effect is driven by anti-robot speciesism.

Third, prior work on speciesism has debated whether derogation of outgroups is driven by real perceived differences between the self versus others, or by a "deeper" speciesism (Horta, 2010; Singer, 1996; Singer & Mason, 2007). The closest test of this has come from comparing humans with severe mental impairments to cognitively more capable monkeys (Caviola et al., 2019). However, in this case there are still several other true differences between humans and monkeys. In contrast, the current studies provide an especially strong demonstration of the power of speciesism, by comparing judgments for humans versus robots that are perfectly humanlike. In doing so, we are among the first to suggest and describe speciesism as it may manifest in interactions with such robots, which one might think of as more akin to actual humans than animals.

**Practical Implications**

Our findings suggest that perfectly humanlike robots will have a mixed reception once they enter our society. While they might be preferred over today's robots, they will continue to face challenges from an adoption and acceptance standpoint, given more entrenched discomfort with them than humans. Our findings suggest that educational interventions explaining their true capabilities might help. Building on intergroup contact theory (Pettigrew & Tropp, 2006), another possibility is that people will overcome anti-robot speciesism if they have enough encounters with such robots, which might lead them to revise stereotypes and alleviate anxiety around interacting with them (Allport, 1954), which might be particularly likely if the robots are framed as being on "your side." Another effective approach may be to align humanoid robots with self-protection (Griskevicius & Kenrick, 2013). For example, since robots lack certain threat cues like disease transmission, which often trigger avoidance behaviors in human interactions, this may be one context in which they are judged more positively than homo sapiens (Ackerman et al., 2018).

Our findings also suggest that how robots are "framed"—whether as general-purpose social entities or as expendable workers in dangerous conditions—can significantly shape human attitudes toward their moral worth, likely affecting support for such services. Of course, if perfectly humanlike robots do truly reach human-level capabilities, then such efforts would give rise to ethical questions concerning our moral obligations toward such robots. If claims of robot consciousness and emotions become difficult to disprove, then denying rights to such robots might inevitably stem from anti-robot speciesism and biological essentialism. Such a denial might persist as a justification for their exploitation.

**Limitations and Future Directions**

We suggest further research on perfectly humanlike robots as we approach their commercialization. First, as robots become more capable, future research can improve realism by exposing people to real humanoid robots. While our own studies could only approximate realistic scenarios by showing participants photos or videos and claiming they depict humanoid robots, we note that participants in our two supplemental studies were already deceived by videos of today's robots. Furthermore, participants in Study 5 found it not harder to imagine such robots than humans. For the same reason, it seems unlikely that our results are explained by familiarity alone, although we cannot rule this out completely. Finally, since participants were more comfortable receiving service from perfectly humanlike robots than from imperfect robots, it also seems unlikely that the results can be explained by fears of replacement (McClure, 2018).

Second, we only examined one direction of technological development: robots becoming more and more human-like. Yet, there is a parallel transformation under way: humans becoming increasingly technologically enhanced, and thus more robotic (Castelo, Schmitt, et al., 2019). As humans become increasingly integrated with technology, will this trend weaken the view that robots are categorically different from humans? Conversely, would future robots made, in part, of biological material, which are currently under development (Truby, 2021), be more acceptable?